%% file: root.tex
\documentclass[letterpaper, 10 pt, conference]{ieeeconf}  

\IEEEoverridecommandlockouts                              

\usepackage{marvosym}
\title{Category-Level 6D Object Pose Estimation via \\ Cascaded Relation and Recurrent Reconstruction Networks}

\author{Jiaze Wang$^{1*}$, Kai Chen$^{1*}$ and Qi Dou$^{1,2}$\textsuperscript{\Letter}
\thanks{$^{1}$ Jiaze Wang, Kai Chen and Qi Dou are with the Department of Computer Science and Engineering at The Chinese University of Hong Kong, Hong Kong SAR, China.
{\tt\small qidou@cuhk.edu.hk}}
\thanks{$^{2}$ Qi Dou is also with T Stone Robotics Institute, The Chinese University of Hong Kong, Hong Kong SAR, China}
\thanks{$^{*}$ Authors contributed equally}
}

\usepackage{subfigure}

\usepackage{times}
\usepackage{epsfig}
\usepackage{graphicx}
\usepackage{amsmath}
\usepackage{amssymb}
\usepackage{multirow}
\usepackage{adjustbox}

\usepackage{etoolbox}
\usepackage{threeparttable}
\makeatletter
\patchcmd{\@makecaption}
  {\scshape}
  {}
  {}
  {}
\makeatletter
\patchcmd{\@makecaption}
  {\\}
  {.\ }
  {}
  {}
\makeatother

\usepackage[pagebackref=true,breaklinks=true,colorlinks,bookmarks=false]{hyperref}

\begin{document}

\maketitle
\thispagestyle{empty}
\pagestyle{empty}

\input{sections/abstract}

\input{sections/introduction}

\input{sections/relatedwork}

\input{sections/methodology}
\input{sections/experiments}

\input{sections/conclusion}
\input{sections/acknowledge}

\input{root.bbl}
\bibliographystyle{IEEEtran}


\end{document}

%% file: sections/abstract.tex
\begin{abstract}





Category-level 6D pose estimation, aiming to predict the location and orientation of unseen object instances, is fundamental to many scenarios such as robotic manipulation and augmented reality, yet still remains unsolved.
Precisely recovering instance 3D model in the canonical space and accurately matching it with the observation is an essential point when estimating 6D pose for unseen objects. In this paper, we achieve accurate category-level 6D pose estimation via cascaded relation and recurrent reconstruction networks.
Specifically, a novel cascaded relation network is dedicated for advanced representation learning to explore the complex and informative relations among instance RGB image, instance point cloud and category shape prior. Furthermore, we design a recurrent reconstruction network for iterative residual refinement to progressively improve the reconstruction and correspondence estimations from coarse to fine. Finally, the instance 6D pose is obtained leveraging the estimated dense correspondences between the instance point cloud and the reconstructed 3D model in the canonical space. We have conducted extensive experiments on two well-acknowledged benchmarks of category-level 6D pose estimation, with significant performance improvement over existing approaches. On the representatively strict evaluation metrics of $3D_{75}$ and $5^{\circ}2 cm$, our method exceeds the latest state-of-the-art SPD~\cite{tian2020shape} by $4.9\%$ and $17.7\%$ on the CAMERA25 dataset, and by $2.7\%$ and $8.5\%$ on the REAL275 dataset. Codes are avaliable at \url{https://wangjiaze.cn/projects/6DPoseEstimation.html}. 

\end{abstract}

%% file: sections/introduction.tex

\section{Introduction}

Accurate 6D pose estimation has increasingly been an important yet challenging research topic in computer vision, which aims to predict the location and orientation of 3D objects~\cite{peng2019pvnet,wang2019densefusion,xiang2017posecnn}.
It has extensive prospects in real-world applications such as robotic manipulation, augmented reality, navigation and 3D scene understanding. In recent years, although pioneering work in \emph{instance-level 6D pose estimation} has made remarkable progress~\cite{he2020pvn3d,li2020robust,song2020hybridpose,tekin2018real}, almost all these methods require exact 3D CAD object models for the instances. However, such an assumption is difficult, if not impossible, to be satisfied in real practice, considering the diversity of object instances as well as the cost for building a CAD model for each instance. In addition, these methods can not handle new instances with unknown CAD models, which impedes the generalizability in environments with previously object instances without CAD models.

\begin{figure}[t]
    \centering
    \includegraphics[width=1.0\linewidth]{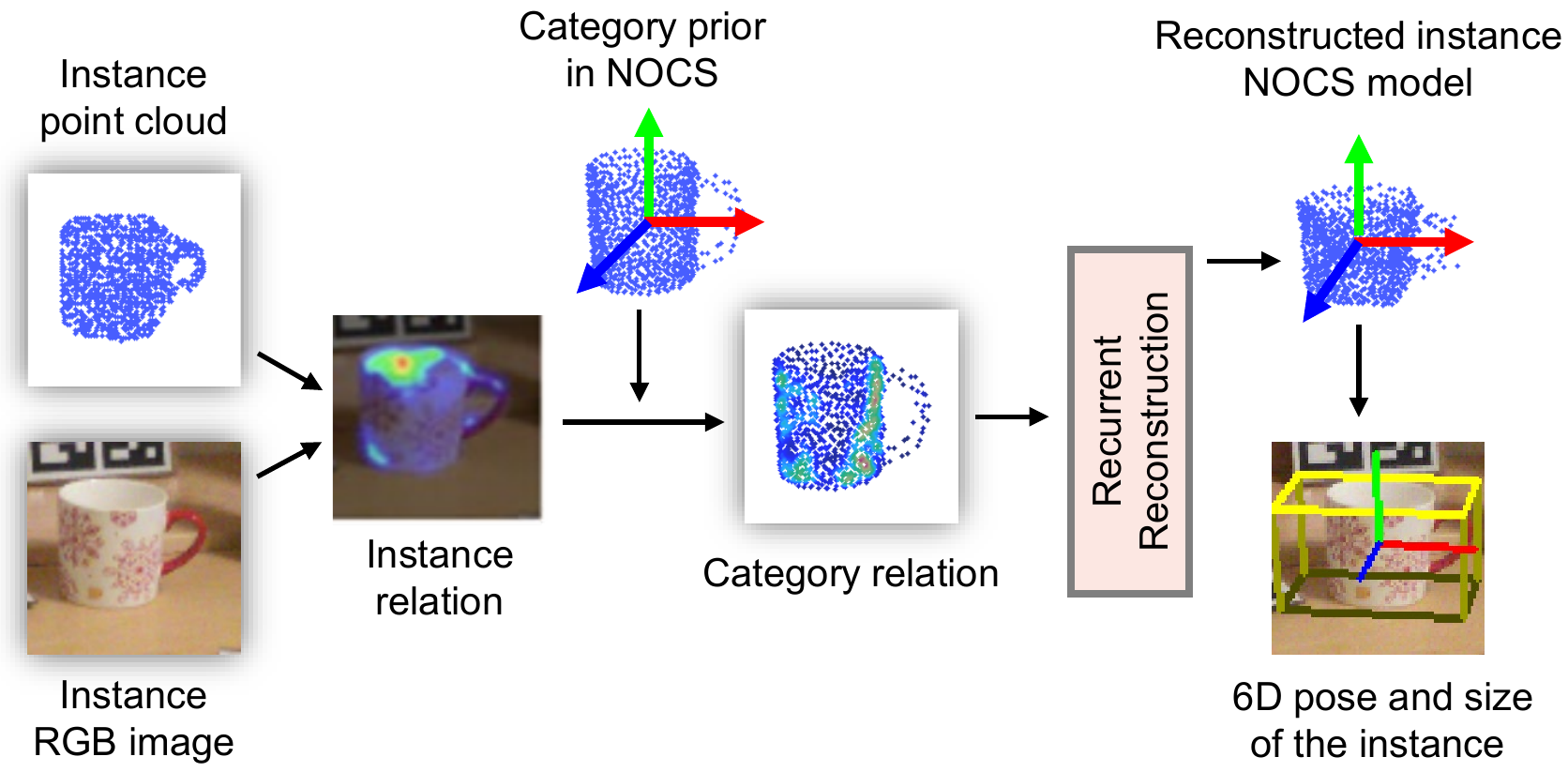}
    \caption{The proposed category-level 6D pose estimation via cascaded relation and recurrent reconstruction networks.}
    \label{fig:coverpage_relations}
    \vspace{-10pt}
\end{figure}


In contrast, the aim of \emph{category-level 6D object pose estimation} is to generate 6D poses for novel object instances of the same category, which is much more challenging. In conventional instance-level methods, object pose is estimated via correspondences between the observed instance RGB or RGB-D image and its exact CAD model. Yet, in category-level setting, such a correspondence can not be directly constructed without a specific CAD model. 
In order to explore such correspondence information in category-level 6D pose estimation, Wang et al.~\cite{wang2019normalized} innovate the Normalized Object Coordinate Space~(NOCS) which is a unified coordinate system. In NOCS, the size of objects are normalized within a defined coordinate space, and the instances belonging to the same category share an identical orientation. Since the object CAD model is unknown, they reconstruct a canonical model representation in NOCS from instance observations, and build dense correspondences between the reconstructed NOCS model with instance image or point cloud for 6D pose estimation. Recently, Tian et al~\cite{tian2020shape} further improve the accuracy of the reconstructed 3D model in NOCS by introducing a shape prior to the reconstruction process. They extract category features from the shape prior and concatenate instance features with the category ones, to address the shape variations across instances within the same category. The quality of reconstructed NOCS model is significantly enhanced by harnessing the shape-based category features.




Though promising progress has been witnessed, it is still extremely challenging to accurately reconstruct the 3D object models in NOCS, which plays a crucial role to boost 6D pose estimation performance.
First, learning representative features from RGB-D image is essential, given that the color image and point cloud data provide complementary texture and geometry information for the objects. Capturing the inherent relations of them improves the representation capability of the instance embeddings. The relevant category feature is also valuable, which helps to model the instance shape variations during reconstruction. Exploiting the relation between such category features with the instance ones helps to overcome intra-class shape variations as well as provide
global contextual clues for NOCS model reconstruction. 
However, such important yet complex relations have not been well investigated in existing methods so far.
Second, reconstructing the object model in the canonical NOCS space forms a dense regression task.
In line with other dense regression problems~\cite{gu2020cascade,hui2018liteflownet}, it is difficult for a network to produce highly accurate regression results with a single step, because the network may partially focus on some object parts while overlooking the others. This hampers the 6D pose estimation accuracy, as the optimization process may be trapped into a local minimum when dealing with spatially biased correspondences.

To address above challenges, we propose a novel method of \textit{Cascaded Relation and Recurrent Reconstruction Networks} for category-level 6D object pose estimation.
As illustrated in Figure \ref{fig:coverpage_relations}, our framework presents cascaded relation networks to capture the informative relations of instance RGB images, instance point cloud, and category priors, which is important for accurate canonical model reconstruction.
Specifically, an instance relation network captures the relation of the input RGB image and point cloud to extract representative feature embeddings for each instance. A category relation network exploits the relation of instance features and category features to address the large shape variations. Moreover, with the relation-enhanced features, a recurrent reconstruction network is then developed to accurately reconstruct the instance model in NOCS space, which progressively refines the reconstructed models from coarse to fine. Finally, we leverage the reconstructed model and the observed point clouds to estimate object 6D pose by point matching.
Our main contributions are summarized as:\vspace{-3pt}



\begin{itemize}
\setlength{\itemsep}{0pt}
    \item We propose a novel cascaded relation network to capture the underlying relations of multi-source inputs. Our network leverages the complementary advantages of these features for categorical object pose estimation.
    \item We design a recurrent reconstruction network to accurately reconstruct the instance 3D model in NOCS space. By iteratively estimating reconstruction residuals, our network progressively refines the model and the correspondence matrix.

    \item We conduct extensive experiments on two well-acknowledged benchmarks with dramatic performance improvement over existing methods. On the representative strict evaluation metrics of $3D_{75}$ and $5^{\circ}2cm$, our method exceeds the latest state-of-the-art SPD~\cite{tian2020shape} by $4.9\%$ and $17.7\%$ on the CAMERA25 dataset, and by $2.7\%$ and $8.5\%$ on the REAL275 dataset. 
    
    
\end{itemize}

%% file: sections/relatedwork.tex
\section{Related Work}


\textbf{Instance-Level 6D Pose Estimation.}
Instance-level 6D object pose estimation methods can be broadly categorized into two categories of RGB-based and RGBD-based methods, according to the format of input data. 
Classical RGB-based methods~\cite{collet2011moped,ferrari2006simultaneous,hinterstoisser2012model} focus on detecting and matching keypoints with known models. Current deep learning methods~\cite{brachmann2014learning,hinterstoisser2011multimodal,li2020robust,tejani2014latent,tekin2018real,tremblay2018deep,huang2020movienet} improve the performance by replacing the hand-crafted keypoint detection and matching process with a data-driven learning scheme that predicts 2D keypoints on RGB images and solves object poses by PnP~\cite{fischler1981random}. Instead of explicitly detecting and matching object keypoints, some methods~\cite{tekin2018real} take corner points of 3D object bounding boxes as keypoints, or implicitly represent keypoints by a dense voting field~\cite{peng2019pvnet}. These methods thus can effectively cope with low-texture environments. Other methods~\cite{kehl2017ssd,li2018deepim,manhardt2018deep} propose to directly predict pose parameters from RGB observations. They extract and group pose-relevant features by CNN and regress translation and rotation with two separate multi-layer perceptrons. Although impressive, these RGB-based methods face problems in complex environments such as cluttering or occlusion~\cite{rad2017bb8,peng2019pvnet}. Lacking depth information increases the ambiguity of estimation and also hurts the pose accuracy. 
For RGBD-based methods, once obtained the intrinsic matrix of a RGB-D camera, we can recover the object point cloud via inverse projection~\cite{he2020pvn3d,wang2019densefusion}. How to make full use of the appearance feature from the RGB image and the complementary geometry feature from point cloud is a major challenge in RGBD-based methods~\cite{song2016deep}. Algorithms such as PoseCNN \cite{xiang2017posecnn} uses them in totally separate steps, in which RGB images are used to predict initial object poses while point clouds are utilized for ICP-based pose refinement \cite{besl1992method}. \cite{li2018unified} fuses the depth image as a new input channel to conventional CNNs which lacks discriminative treatments for point clouds. Recent methods \cite{he2020pvn3d,song2020hybridpose,wang2019densefusion} densely fuse the color embedding with the geometry embedding in a pixel-wise manner. Unfortunately, they still did not consider any global feature relations~\cite{wang2020cascade,rao2020unified,rao2020local,wang2019part,xia2020online} during fusion. 

\textbf{Category-Level 6D Pose Estimation.}
Existing works on category-level 6D pose estimation are still scarce to date.
Compared with instance-level object pose estimation, the category-level task is more challenging due to the large intra-class variations in aspects of texture and shape among instances. Establishing an intermediate representation that reduces such difference is a widely applied idea in existing algorithms. Sahin et al.~\cite{sahin2018category} divide an object into multiple 3D skeleton structures, from which they derive a shape-invariant representation and develop a part-based random forest architecture for categorical 6D pose estimation. By integrating 3D shape estimates from a generative object model, \cite{burchfiel2019probabilistic} produces a distribution over predicted poses with only rotation information. 
Alternative methods~\cite{wang2019normalized} express an object in canonical coordinate space. By inferring / regressing the object canonical representation and associate it with the specific instance observation, 6D object pose can be determined without 3D CAD models. 
Wang et al.~\cite{wang2019normalized} introduce the Normalized Object Coordinate Space to represent different object instances within a category in a unified manner. Then a network is trained to predict correspondences from object pixels to points in NOCS. Subsequently, these correspondences are used with the depth map to estimate 6D pose and size by point matching. Inspired by NOCS, the first category-level pose tracker is proposed by \cite{wang20206}. Recent methods~\cite{chen2020learning,tian2020shape} propose to leverage the category-related features to explicitly model the shape variation when reconstructing the canonical representation. 
In this paper, we explore the complex relations among the texture feature, geometry feature and category feature, in addition to progressively recover the object canonical model with the enhanced feature representations.

%% file: sections/methodology.tex
\section{Method}
\begin{figure*}[ht]
	\centering
	\includegraphics[width=0.9\linewidth]{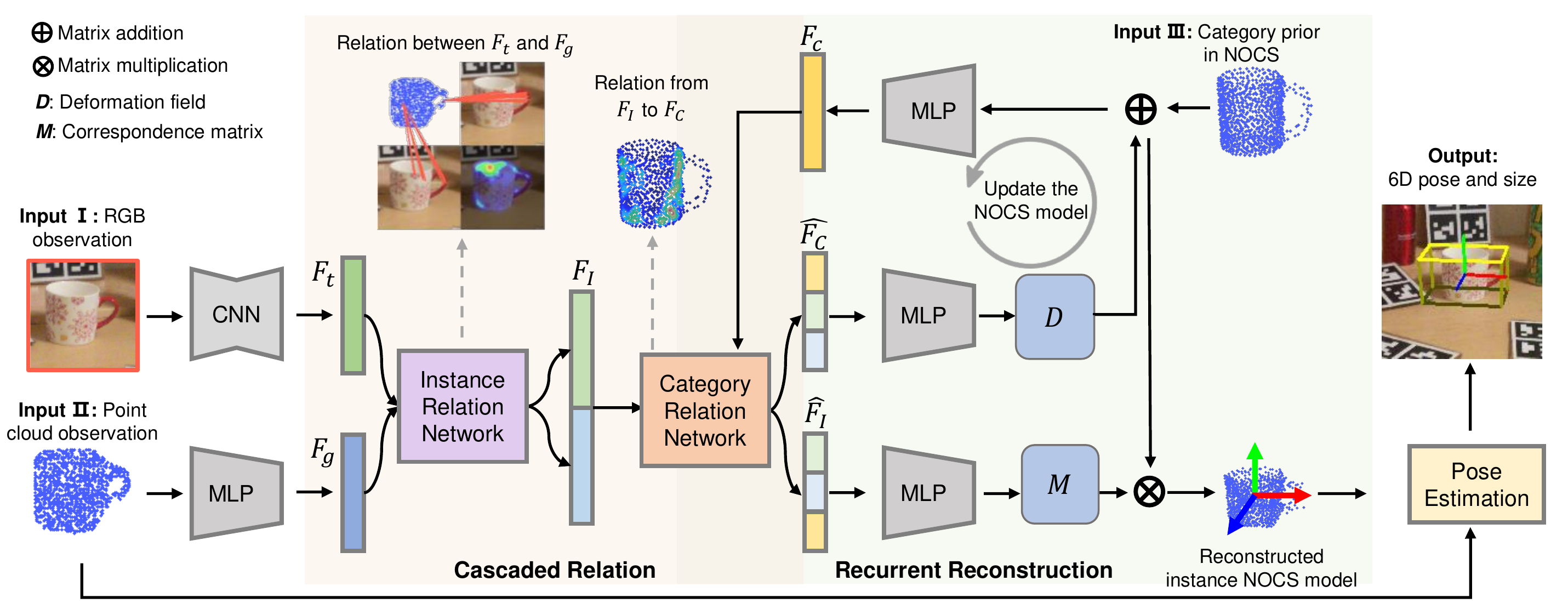}
	\caption{Overview of our Cascaded Relation and Recurrent Reconstruction Networks. The networks are mainly composed of two networks: (1) A cascaded relation network to exploit the relation of between RGB images and point clouds, and the relation between instance features and category features. (2) An recurrent reconstruction network for canonical shape reconstruction from coarse to fine.}
	\label{fig:framework}
    \vspace{-5pt}
\end{figure*} 

In this section, to make the contents self-contained and easy to follow, we will first introduce the useful preliminary, then briefly describe an overview of our proposed category-level 6D pose estimation framework. Next, we will describe the proposed novel cascaded relation network, recurrent reconstruction network and pose generation method in detail.

\subsection{Preliminary}\label{subsec:preliminary}
Given a calibrated RGB-D image, we aim to estimate a 6D pose for an object of interest, which is a rigid transformation $[R|t]$ composed of a rotation $R\in SO(3)$ and a translation $t\in \mathcal{R}^3$ components. To process scenes containing multiple instances with different categories, we first employ an off-the-shelf instance segmentation network~(i.e., Mask-RCNN~\cite{he2017mask}) to detect and segment each individual object instance. The yielded detection bounding box is used to crop the RGB image into object patches, and the segmentation mask is leveraged to convert the depth image into object point cloud~(through camera intrinsic matrix). Inspired by~\cite{tian2020shape,wang2019normalized}, once we can reconstruct the exact instance 3D model in the NOCS canonical space, the problem of pose estimation then is reduced to determining the similarity transformation from instance point cloud to the reconstructed 3D model. In order to integrate the predicted object category information into this scheme, we further build an initial 3D canonical model for each category of objects and take it as a category-level prior feature.

\subsection{Framework Overview}\label{subsec:overview}
As shown in Figure~\ref{fig:framework}, our novel pose estimation framework has three inputs: instance RGB image, instance point cloud and the corresponding category prior in NOCS. Estimating object 6D poses with the above multi-source inputs, Wang et al.~\cite{wang2019normalized} separately process them with independent network modules, while Tian et al.~\cite{tian2020shape} integrating them by concatenating their features in latent space.
In contrast, we aim to devise an effective strategy to fully leverage complementary knowledge among the provided inputs for pose estimation. Particularly, we propose a cascaded relation network that constructs the relationship context for input features in two cascaded stages. The first stage models relations between instance image feature and point cloud feature to obtain representative instance features. The second stage further correlates the instance features with category features. 
On top of these, a dense deformation field is regressed for the use of adjusting the category prior for NOCS model reconstruction. Meanwhile, a correspondence matrix is also estimated to match the instance point cloud with the reconstructed model. Furthermore, we develop a recurrent reconstruction network for accurate reconstruction and matching. The network iteratively updates the reconstructed model and the category-level feature in each recurrent step. By exploiting multi-stage supervisions, it learns residuals to progressively refine the deformation field and the correspondence matrix.
Finally, given the instance NOCS model and the correspondence matrix, object 6D pose and size can be generated by correspondence-based optimization.




\subsection{Cascaded Relation Network}\label{subsec:cascaded_relation}
In the following, we describe the proposed cascaded relation network. Formally, we denote an instance observation by $(I,V)$, with $I\in \mathcal{R}^{H \times W \times 3}$ being the image patch, and $V\in \mathcal{R}^{N_p \times 3}$ being the point cloud recovered from the depth map. $N_p$ is the number of points in $V$. Let $V_c\in \mathcal{R}^{N_c\times 3}$ be the corresponding category prior where $N_c$ denotes the number of points in $V_c$. We first resort to CNN and MLPs to extract texture and geometry features from $I$, $V$ and $V_c$, respectively. After that, similar to~\cite{wang2019densefusion}, we align the texture feature map to point clouds and associate each instance point with a texture feature vector. Consequently, we get the instance texture feature $F_t\in \mathcal{R}^{C_t \times N_p}$, the instance geometry feature $F_g\in \mathcal{R}^{C_g \times N_p}$, and the category feature $F_c\in \mathcal{R}^{C_c\times N_c}$, where $C_\ast$ is the feature channels.

The relations among $F_t$, $F_g$ and $F_c$ are important for reconstructing the NOCS model towards accurate pose estimation for an instance.
On the one hand, it is supposed to capture the characteristics of the instance so that the reconstructed model can well match the observed point clouds. On the other hand, the reconstruction network should also account for certain general attributes of the category, so that the reconstruction process can overcome intra-class variations thus enhancing generalizability to unseen instances of the same category. In these regards, we propose a cascaded relation network to harness these two kinds of relations for accurate NOCS reconstruction and pose estimation.

\textbf{Instance Relation.} The instance relation network~(IRN) is designed to learn the complementary knowledge between $F_t$ and $F_g$, with $F_t$ encoding instance texture and semantic features and $F_g$ encoding its geometry information. These two features complement each other in key aspects. For example, due to absence of depth information, $F_t$ extracted by CNN is susceptible to image background and cluttering environments, which can be alleviated in $F_g$ from point cloud. The MLP produces $F_g$ from disordered point clouds using a narrow receptive field, leading to less efficacy on spatial-aware representations, which in turn can be mitigated with $F_t$. Given these properties, we propose to capture their relations and deeply integrate them using a relation function $\mathcal{G}^i$. In this way, we reproduce the relation-injected texture feature and geometry feature of the instance as follows:
\begin{equation}
    \hat{F_t} = F_t + \mathcal{G}^i(F_t, F_g), \quad
    \hat{F_g} = F_g + \mathcal{G}^i(F_g, F_t).
\end{equation}
We add the original feature to the relation feature, in order to drive the network to unearth as much complementary information as possible by relational learning. We concatenate $\hat{F_t}$ and $\hat{F_g}$ as our relation-enhanced instance embedding $F_I$, which is expected to represent the instance's characteristics in the camera frame.

\textbf{Category Relation.} Next, the category relation network (CRN) aims to capture relations between the $F_I$ and $F_c$
that presents the category-level feature of an instance in NOCS space. The interaction between these features helps the network to accurately model the shape variation of instances from the same category. We cascade CRN behind IRN, with the careful consideration that the strong relation-enhanced instance embedding $F_I$ should be yielded beforehand.
Directly associating those original $F_t$ and $F_g$ with $F_c$ may cannot fully tap the potential of relational learning in category-level. Similar to our IRN, we initiate a relation function $\mathcal{G}^c$ to exploit relations between $F_I$ and $F_c$:
\begin{equation}
    \hat{F_I} = F_I + \mathcal{G}^c(F_I, F_c), \quad
    \hat{F_c} = F_c + \mathcal{G}^c(F_c, F_I).\vspace{-3pt}
\end{equation}
The relation-injected features $\hat{F_I}$ and $\hat{F_c}$ then will be used in the subsequent model reconstruction and pose estimation.

\textbf{Choice of Relation Function $\mathcal{G}$}. The formulation of our cascaded relation network is flexible, which is not restricted by the specific architecture design for $\mathcal{G}$. Any structure capable of capturing feature relations can be easily adopted as $\mathcal{G}$ in our framework. In this paper, we employ three representative and popular structures: MLP~\cite{santoro2017simple}, Non-local~\cite{wang2018non}, and Transformer~\cite{devlin2018bert}. The experimental results indicate that the cascaded relation network improves the pose estimation performance significantly, no matter which specific $\mathcal{G}$ is utilized. Please refer to Table~\ref{table:relation} for comparison results.

\subsection{Recurrent Reconstruction Network}\label{subsec:recurrent_reconstruction}
In this section, we describe the recurrent reconstruction network.
With the $\hat{F_I}$ and $\hat{F_c}$ produced by CRN, as in~\cite{tian2020shape}, we regress a deformation $D\in \mathcal{R}^{N_c\times 3}$ from the initial category model to the instance canonical model that we aim to reconstruct. Meanwhile, we regress a correspondence matrix $M\in \mathcal{R}^{N_p\times N_c}$ for associating the instance point cloud with the reconstructed 3D model in NOCS. Each row of $M$ indicates a weighted correspondence between a point in $V$ and all points in $V_c$. Even with our relation-enhanced features, accurately regressing all these values within one single step is still infeasible. The points at different locations may have diverse deformations. The network may have bias towards some locations but neglect the remaining points, which results in an inaccurate reconstruction model. Accordingly, the correspondence matrix will also be adversely affected.

To address this problem, we propose a recurrent reconstruction network to regress $D$ and $M$ from coarse to fine. Our network first predicts an initial deformation field $D^0$ and an initial correspondence matrix $M^0$ directly from the relation-enhanced features $\hat{F_c}$ and $\hat{F_I}$. Then, taking $D^0$ and $M^0$ as initial values, we estimate the residuals to refine the deformation field and correspondence matrix. Specifically, we add $D^0$ to $V_c$ to update the NOCS model, and denote its updated category features as $F_c^1$. After that, we integrate $F_c^1$ with $F_I$ in the CRN to model the shape variation for the recurrent step and make the current reconstruction focus more on regions that are neglected in previous steps. For the first recurrent step, our network would target on estimating the deformation residual $\bar{D}^1=D_{gt} - D^0$ and correspondence residual $\bar{M}^1=(M^0)^{-1}\times M_{gt}$. To further improve the accuracy, we can repeat the recurrent optimization multiple times. For the $i$-th recurrent step, its deformation and correspondence matrix can be computed as:
\begin{equation}
    D^{i} = D^{i-1} + \bar{D}^i, \quad
    M^{i} = M^{i-1}\times \bar{M}^i.
\end{equation}
The outputs of the last recurrent step are the final estimation results for the deformation field and correspondence matrix.

Notably, the loss functions to supervise the recurrent reconstruction process are crucial.
We impose supervisions on $D$ and $M$ to drive the network learning to reconstruct the instance NOCS model and associate it with the observed point clouds accurately.

\textbf{Reconstruction Loss.} Applying the deformation field $D$ on $V_c$ yields the reconstructed NOCS model $R$. During training, with the ground-truth NOCS model $R_{gt}$ for each instance, we employ the reconstruction loss to penalize $D$. Specifically, we exploit the Chamfer distance~(CD) to measure the similarity between $R$ and $R_{gt}$ as:
\begin{equation}
\setlength{\belowdisplayskip}{1pt}
L_{r}=\sum_{i\in R^{i}}\underset{j\in R_{gt}^{j}}{min}\left \| i-j \right \|_{2}^{2}+\sum_{j\in R_{gt}^{j}}\underset{i\in R^{i}}{min}\left \| i-j \right \|_{2}^{2}.
\end{equation}
In addition, a regularization loss is further added to penalize large deformations: $L_\text{def}=\frac{1}{N_c}\sum_{i\in D }\left \| i \right \|_{2}$.

\textbf{Correspondence Loss.} We supervise $M$ with a correspondence loss function inspired by~\cite{tian2020shape}. After applying $M$ on the reconstructed NOCS model, we can get the NOCS coordinate prediction for each point in $V$. Again, since the ground-truth NOCS coordinate for the point cloud observation is known during training, we supervise $M$ by constraining the distance between the predicted NOCS coordinate value $x$ and the ground-truth one $x_{gt}$. The correspondence loss $L_{o}$ can be defined as:
\begin{equation}
\setlength{\belowdisplayskip}{1pt}
L_{o}(x,x_{gt})= \frac{1}{N_c}\left\{\begin{matrix}
5(x-x_{gt})^{2} & \left | x-x_{gt} \right | \leq 0.1,\\ 
\left | x-x_{gt} \right | -0.05 &  \text{otherwise},
\end{matrix}\right.
\end{equation}
in which a soft $L_{1}$ loss is used for robust optimization. In addition, the same regularization loss $L_\text{reg}$ as the one in~\cite{tian2020shape} is further adopted to constrain the sparsity of $M$.


\textbf{Recurrent Loss.} We combine the above two loss terms in one recurrent step. Moreover, we exploit deep supervision mechanism~\cite{deng2018r3net} that imposes supervision on every recurrent step. The supervision on intermediate results impels the network to learn the residuals from the ground truth, and the accumulated loss function is as:
\begin{equation}
\setlength{\belowdisplayskip}{1pt}
    L_\text{overall}=\sum_{k=0}^{N} \lambda_{k} \times (L^k_r + L^k_\text{def} + L^k_o + L^k_\text{reg}),
\end{equation}
where $L_r^k,L^K_\text{def},L_o^k,L_\text{reg}^k$ denotes the loss for the $k$-th recurrent step and $\lambda_{k}$ is the associated weighted factor. This overall loss is used to train the recurrent reconstruction network.

\subsection{Correspondence based Pose Estimation}\label{subsec:pose_estimation}
Feeding the instance RGB image $I$, instance point cloud $V$, and the category prior $V_c$ into the framework, our network would output a deformation field $\hat{D}$ and a correspondence Matrix $\hat{M}$. As in~\cite{tian2020shape,wang2019normalized}, we can use $\hat{D}$ and $\hat{M}$ to estimate the specific 6D object pose. First of all, we apply $\hat{D}$ on $V_c$ to get the reconstructed NOCS model $V_{nocs}$ for the instance. Note that $V_{nocs}$ is in the canonical NOCS space, and the point cloud observation $V$ is in the camera coordinate space. Once we recover $V_{nocs}$, estimating the 6D pose for the instance in the camera frame would be equal to finding a similarity transformation from $V_{nocs}$ to $V$ up to a scale factor. This similarity transformation can be estimated based on dense correspondences between $V_{nocs}$ and $V$. To obtain these correspondences, we apply $\hat{M}$ on $V_{nocs}$ and compute a NOCS coordinate for each point in $V$. Finally, we use the Umeyama algorithm~\cite{umeyama1991least} to estimate the similarity transformation, in which the 3D rotation and translation corresponds to the 6D object pose, and the scale factor corresponds to the object size. The RANSAC~\cite{fischler1981random} is also adopted to remove outliers and achieve a robust estimation.

%% file: sections/experiments.tex
\section{Experiments}
\begin{table*}[!ht]
\centering
\caption{Comparison of our method with current state-of-the-art methods on both benchmarks.} \vspace{-5pt}
\label{table:stoa}
\resizebox{\textwidth}{!}{%
\begin{threeparttable}
\begin{tabular}{c|cccccc|cccccc}
\hline
\multirow{2}{*}{Method} & \multicolumn{6}{c|}{CAMERA25} & \multicolumn{6}{c}{REAL275} \\ \cline{2-13} 
 & $3D_{50}$ & $3D_{75}$ & $5^{\circ}2cm$ & $5^{\circ}5cm$ & $10^{\circ}2cm$ & $10^{\circ}5cm$ & $3D_{50}$ & $3D_{75}$ & $5^{\circ}2cm$ & \multicolumn{1}{c}{$5^{\circ}5cm$} & $10^{\circ}2cm$ & $10^{\circ}5cm$ \\ \hline
NOCS \cite{wang2019normalized} & 83.9 & 69.5 & 32.3 & 40.9 & 48.2 & 64.6 & 78 & 30.1 & 7.2 & \text{10.0} & 13.8 & 25.2 \\
SPD \cite{tian2020shape} & 93.2 & 83.1 & 54.3 & 59.0 & 73.3 & 81.5 & 77.3 & 53.2 & 19.3 & 21.4 & 43.2 & 54.1 \\
CASS\cite{chencategory} & - & - & - & - & - & - & 77.7 & - & - & 23.5 & - & 58.0 \\
SPD\tnote{*} & 93.5 & 87.0 & 59.7 & 64.4 & 77.4 & 84.8 & 77.5 & 53.8 & 20.2 & 22.9 & 42.5 & 52.1 \\
Ours & \textbf{93.8} & \textbf{88.0} & \textbf{72.0} & \textbf{76.4} & \textbf{81.0} & \textbf{87.7} & \textbf{79.3} & \textbf{55.9} & \textbf{27.8} & \textbf{34.3} & \textbf{47.2} & \textbf{60.8} \\ \hline
\end{tabular}%
 \begin{tablenotes}
        \footnotesize
        \item[*] SPD implemented by ourselves with the open source codes.
      \end{tablenotes}
    \end{threeparttable}
}
\end{table*}\vspace{-2pt}

\begin{table*}[!ht]
\centering
\caption{Evaluation of NOCS model reconstruction results regarding the shape quality using the Chamfer Distance (CD) metric ($\times 10^{-3}$).}\vspace{-8pt}
\label{table:reconstruction}
\resizebox{\textwidth}{!}{%
\begin{tabular}{c|ccccccc|ccccccc}
\hline
\multirow{2}{*}{Method} & \multicolumn{7}{c|}{CAMERA25} & \multicolumn{7}{c}{REAL275} \\ \cline{2-15} 
 & Bottle & Bowl & Camera & Can & Laptop & Mug & Average & Bottle & Bowl & Camera & Can & Laptop & Mug & Average \\ \hline
SPD \cite{tian2020shape} & 1.81 & 1.63 & 4.02 & 0.97 & 1.98 & 1.42 & 1.97 & 3.44 & 1.21 & 8.89 & 1.56 & 2.91 & \textbf{1.02} & 3.17 \\
Ours & \textbf{1.30} & \textbf{0.95} & \textbf{2.58} & \textbf{0.82} & \textbf{0.87} & \textbf{1.05} & \textbf{1.18} & \textbf{2.99} & \textbf{0.96} & \textbf{7.57} & \textbf{1.31} & \textbf{1.25} & 1.19 & \textbf{2.53} \\ \hline
\end{tabular}%
}
\end{table*}

\begin{figure*}[!ht]
    \centering
    \begin{minipage}{1.0\textwidth}
        \begin{minipage}{0.03\textwidth}
            \begin{center}
                \rotatebox{90}{SPD}
            \end{center}
        \end{minipage}
    \begin{minipage}{0.96\textwidth}
        \includegraphics[width=0.162\linewidth]{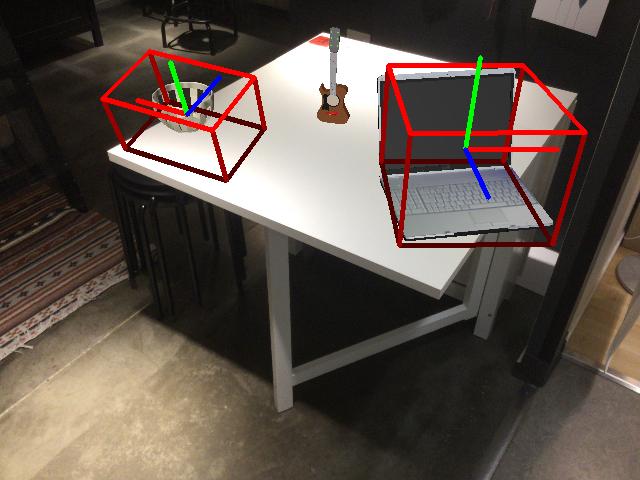}
        \includegraphics[width=0.162\linewidth]{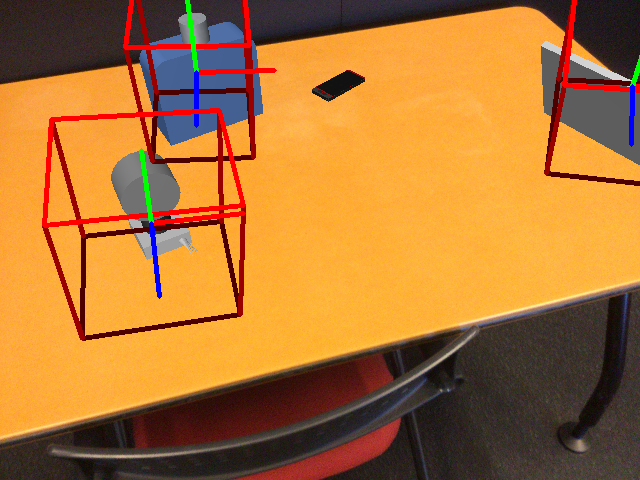}
        \includegraphics[width=0.162\linewidth]{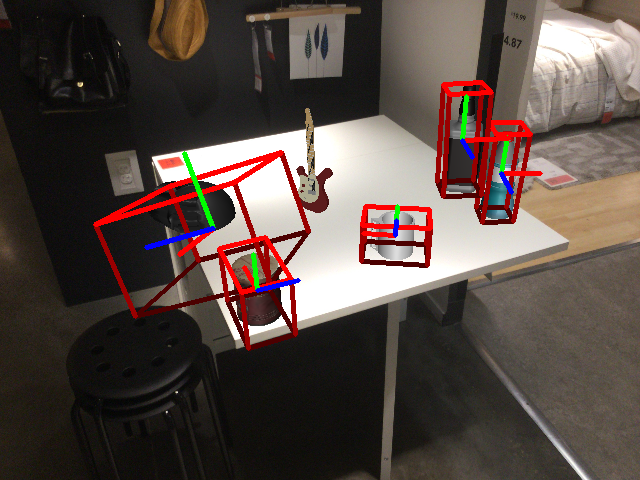}
        \includegraphics[width=0.162\linewidth]{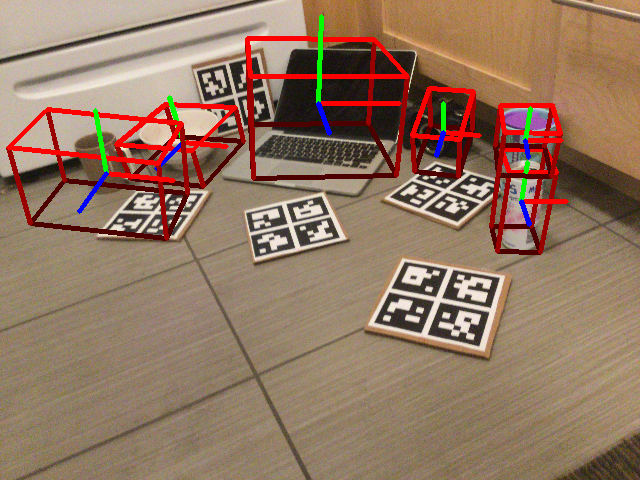}
        \includegraphics[width=0.162\linewidth]{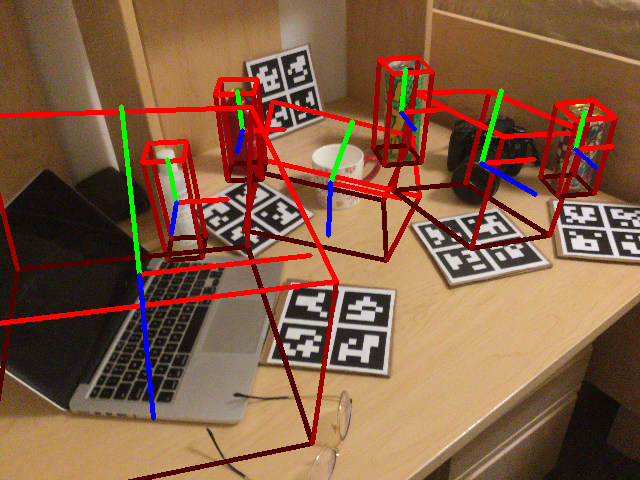}
        \includegraphics[width=0.162\linewidth]{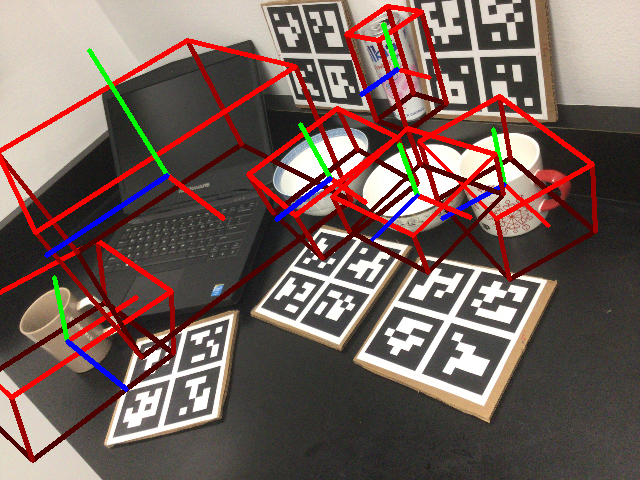}
    \end{minipage}
    \end{minipage}
    \vfill
    \begin{minipage}{1.0\textwidth}
    \begin{minipage}{0.03\textwidth}
        \begin{center}
        \rotatebox{90}{Ours}
        \end{center}
    \end{minipage}
    \begin{minipage}{0.96\textwidth}
        \includegraphics[width=0.162\linewidth]{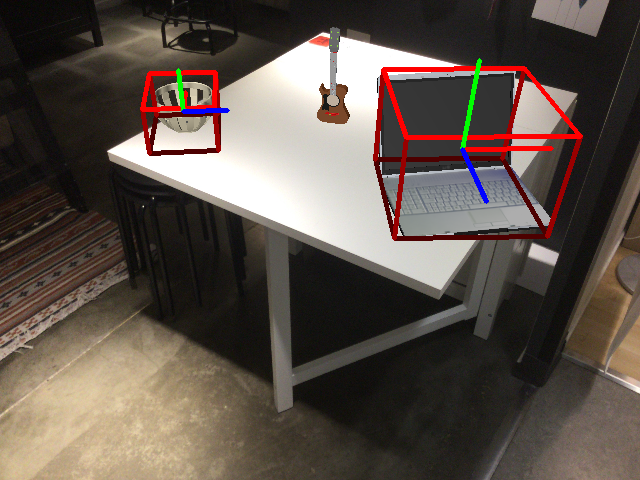}
        \includegraphics[width=0.162\linewidth]{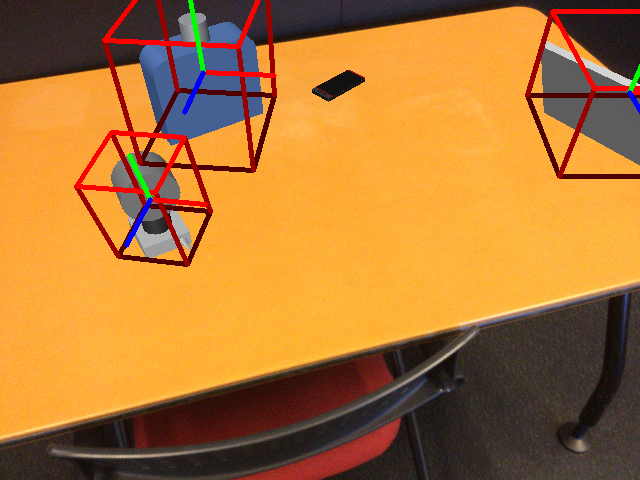}
        \includegraphics[width=0.162\linewidth]{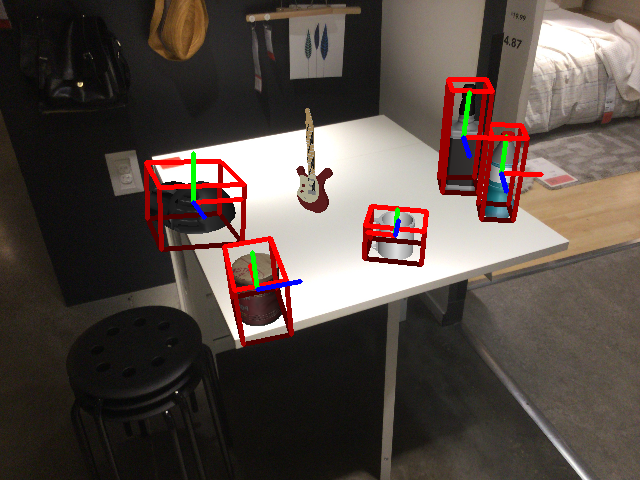}
        \includegraphics[width=0.162\linewidth]{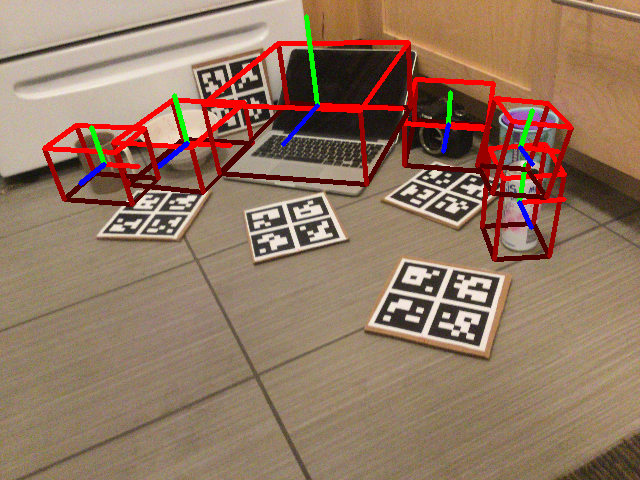}
        \includegraphics[width=0.162\linewidth]{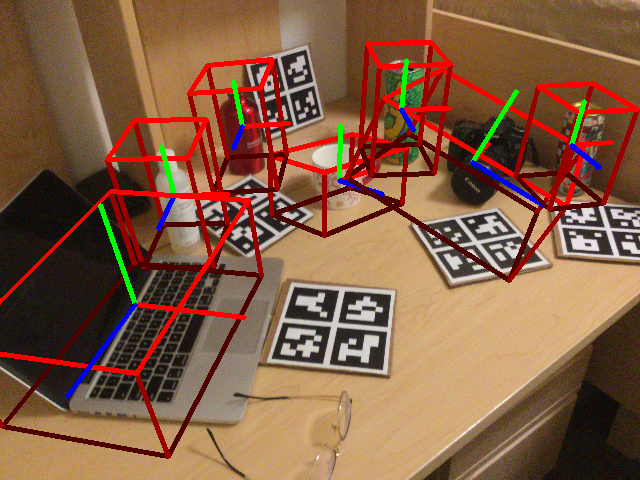}
        \includegraphics[width=0.162\linewidth]{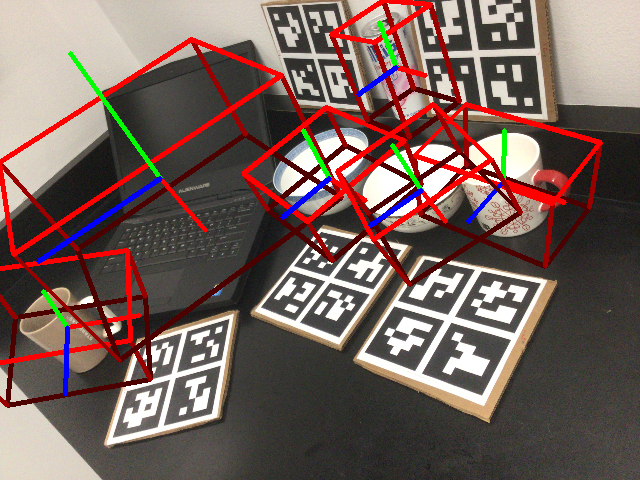}
    \end{minipage}
    \end{minipage}
    \vspace{-2pt}
    \caption{Visual comparison of 6D object pose and size estimation results by our method and SPD~\cite{tian2020shape}. In each row, the first three images are from CAMERA25, the remaining three images are from REAL275.}\
    \label{fig:qua_com}
    \vspace{-8pt}
\end{figure*}

We conduct extensive experiments on two state-of-the-art benchmark datasets of category-level 6D pose estimation, i.e., CAMERA25 and REAL275. 
We have compared with the existing methods on this task, and outperform them by a significant margin on both datasets.
We have also presented the ablation studies to analyze the individual behavior of our proposed cascaded relation network and recurrent reconstruction network.

\subsection{Experimental Setup}
\textbf{Datasets.} We conducted our experiments following the settings in NOCS~\cite{wang2019normalized}. It consists of two datasets: the CAMERA25 dataset and the REAL275 dataset, which corresponds to virtual and real environments respectively. Specifically, CAMERA25 contains 300K RGB-D images (with 25K for evaluation) which are generated by rendering and compositing synthetic objects into real scenes.
REAL275 contains 4300 real-world RGB-D images from 7 scenes for training, and 2750 real-world RGB-D images from 6 scenes for evaluation (with 3 instances per category).
Both datasets contain the same six categories, i.e., \emph{bottle, bowl, camera, can, laptop} and \emph{mug}.


\textbf{Evaluation Metrics.} Similar to~\cite{chen2020learrotationng,tian2020shape,wang2019normalized}, we quantitatively evaluate the estimated 6D object pose on the following metrics: \textbf{3D IoU}: It measures the accuracy of the predicted 3D object bounding box. The predicted pose and the ground truth pose can determine two 3D bounding boxes. Only when the overlapping of these two boxes is larger than a predefined threshold, the predicted pose is judged to be correct. In our experiment, we use $3D_{50}$ and $3D_{75}$, which take $50\%$ and $75\%$ as the Intersection over Union~(IoU) threshold respectively.
\textbf{$\mathbf{a^{\circ}b}$ \textbf{cm}}: It measures the error of predicted poses. Only when the rotation error is less than $\mathbf{a}^{\circ}$ and the translation error is less than $\mathbf{b}$ cm, the pose is judged to be correct. Similar to~\cite{tian2020shape}, we test four different settings: $5^{\circ}2$ cm, $5^{\circ}5$ cm, $10^{\circ}2$ cm, and $10^{\circ}5$ cm.
The rotation error of vertical axis is ignored for symmetrical object categories (\emph{bottle, bowl} and \emph{can}). Similar to~\cite{tian2020shape}, for \emph{mug} category, we treat it as symmetrical object when the handle is not visible, otherwise as asymmetric object.
In addition, the Chamfer Distance(CD) is used to evaluate the instance NOCS model reconstruction accuracy.


\textbf{Implementation Details.}
The texture embedding network is PSPNet \cite{zhao2017pyramid} with backbone of ResNet-18 \cite{he2016deep}, and the model is initialized with pre-trained models from ImageNet \cite{deng2009imagenet}. The image crop is resized to 192 $\times$ 192. The number of points in the input point cloud and category prior is downsampled to 1024. We develop our baseline based on the structure of SPD \cite{tian2020shape}. We train our networks for 50 epochs in total, and there are 4K iterations in each epoch. We use the ADAM\cite{kingma2014adam} to train the network, where the initial learning rate is set as 1e-4 and with weight decay of 1e-6. The learning rate is decreased by a factor of 10 for every 10 epochs. 
Our framework is implemented with PyTorch using 4 TITAN Xp GPUs.
Section~\ref{subsec:sota_experiment} reports our results with a Transformer-based relation network and a recurrent reconstruction network with fixed 2 recurrent stages. The relevant ablation study is presented in Section~\ref{subsec:ablation}. Table \ref{table:relation} focuses on cascaded relation network configurations without adding the recurrent reconstruction module. Similarly, Table \ref{table:recurrent} studies recurrent steps without involving cascaded relations.

\begin{table*}[!ht]
\centering
\caption{Evaluation of Cascaded Relation Network. The ``-", ``M", ``N", and ``T" refer to ``Without relation network", ``MLP based relation network", ``Non-Local based relation network", and ``Transformer based relation network", respectively. For example, ``T / -" denotes the instance relation network is Transformer and the category relation network is none. We report the mAP for the six different metrics.}\vspace{-10pt}
\label{table:relation}
\resizebox{\textwidth}{!}{%
\begin{tabular}{c|cccccc|cccccc}
\hline
\multirow{2}{*}{$\mathcal{G}$} & \multicolumn{6}{c|}{CAMERA25} & \multicolumn{6}{c}{REAL275} \\ \cline{2-13} 
 & $3D_{50}$ & $3D_{75}$ & $5^{\circ}2cm$ & $5^{\circ}5cm$ & $10^{\circ}2cm$ & $10^{\circ}5cm$ & $3D_{50}$ & $3D_{75}$ & $5^{\circ}2cm$ & $5^{\circ}5cm$ & $10^{\circ}2cm$ & $10^{\circ}5cm$ \\ \hline
- / - & 93.5 & 87.0 & 59.7 & 64.4 & 77.4 & 84.8 & 77.5 & 53.8 & 20.2 & 22.9 & 42.5 & 52.1 \\
M / M & 93.3 & 87.1 & 60.4 & 68.2 & 79.4 & 86.0 & 77.1 & 51.1 & 23.9 & 29.6 & 46.2 & 59.3 \\
N / N & 93.2 & 87.0 & 65.6 & 70.2 & 80.4 & 86.7 & 78.0 & 54.5 & 26.3 & \textbf{30.4} & 45.7 & 58.7 \\
- / T & 93.3 & 87 & 62.4 & 68.5 & 78.5 & 84.6 & 78.5 & 54.4 & 23.8 & 28.5 & 45.5 & 57.7 \\
T / - & \textbf{94.5} & \textbf{87.3} & 62.3 & 66.8 & 78.3 & 85.5 & 77.3 & \textbf{57.4} & 25.3 & 28.6 & 46.4 & 58.7 \\ 
T / T & 94.3 & 87.1 & \textbf{70.9} & \textbf{75.7} & \textbf{80.5} & \textbf{87.5} & \textbf{78.7} & 55.8 & \textbf{26.5} & \textbf{30.4} & \textbf{46.5} & \textbf{59.6} \\\hline
\end{tabular}%
}
\vspace{-0.2cm}
\end{table*}

\begin{table*}[!ht]
\vspace{3mm}
\centering
\caption{Evaluation of Recurrent Reconstruction Network. We report the mAP for 6 different metrics on object pose estimation.}\vspace{-5pt}
\label{table:recurrent}
\resizebox{\textwidth}{!}{%
\begin{tabular}{c|cccccc|cccccc}
\hline
\multirow{2}{*}{Step} & \multicolumn{6}{c|}{CAMERA25} & \multicolumn{6}{c}{REAL275} \\ \cline{2-13} 
 & $3D_{50}$ & $3D_{75}$ & $5^{\circ}2cm$ & $5^{\circ}5cm$ & $10^{\circ}2cm$ & $10^{\circ}5cm$ & $3D_{50}$ & $3D_{75}$ & $5^{\circ}2cm$ & $5^{\circ}5cm$ & $10^{\circ}2cm$ & $10^{\circ}5cm$ \\ \hline
0 & 93.5 & 87.0 & 59.7 & 64.4 & 77.4 & 84.8 & 77.5 & 53.8 & 20.2 & 22.9 & 42.5 & 52.1 \\
1 & \textbf{93.7} & 87.8 & 62.2 & 66.8 & 78.4 & \textbf{85.7} & 77.5 & 55.6 & 24.2 & 25.6 & 46.3 & 60.1 \\
2 & 93.6 & 88.6 & \textbf{64.4} & \textbf{68.5} & \textbf{78.9} & 85.6 & \textbf{81.9} & \textbf{57.0} & 24.2 & 25.7 & \textbf{46.4} & \textbf{60.2} \\
3 & 93.6 & \textbf{88.7} & 64.3 & 68.3 & 78.8 & 85.6 & 81.5 & 56.8 & \textbf{24.5} & \textbf{26.2} & 46.0 & 59.5 \\ \hline
\end{tabular}%
}
\vspace{-0.3cm}
\end{table*}

\begin{figure*}[!ht]
    \centering
    \begin{minipage}{0.95\textwidth}
        \includegraphics[width=0.245\linewidth]{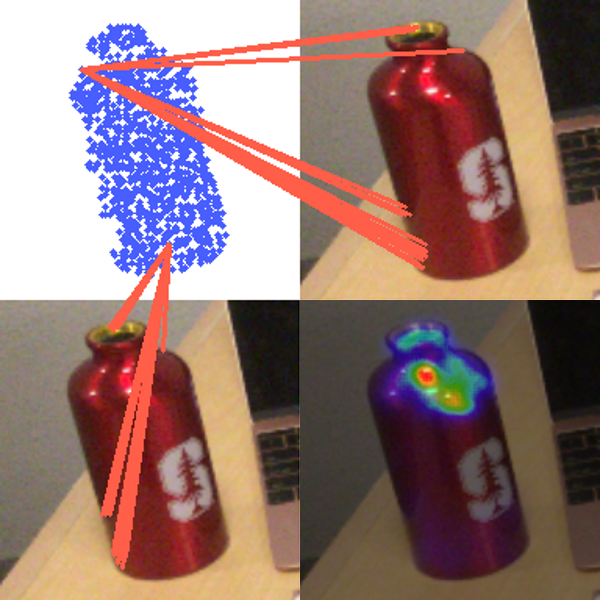}
        \includegraphics[width=0.245\linewidth]{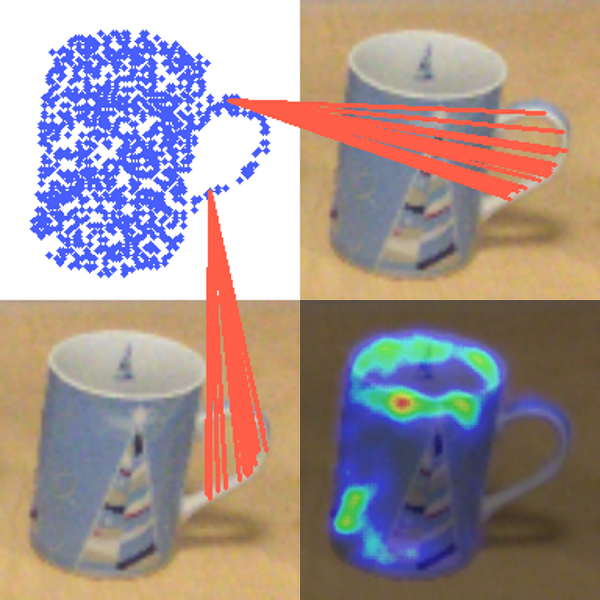}
        \includegraphics[width=0.245\linewidth]{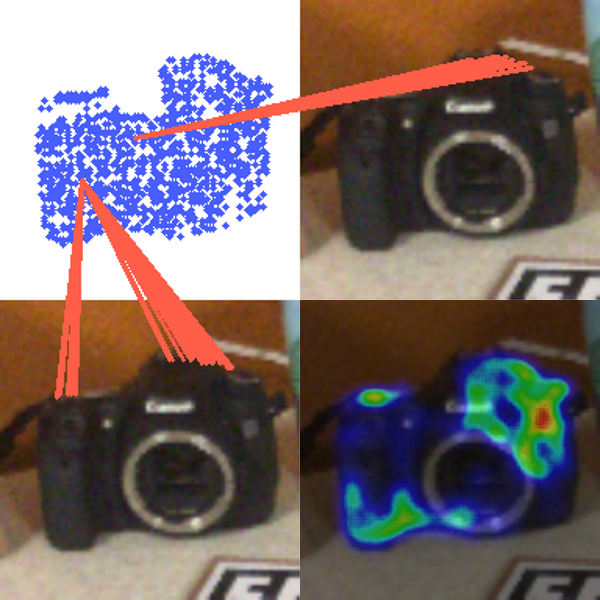}
        \includegraphics[width=0.245\linewidth]{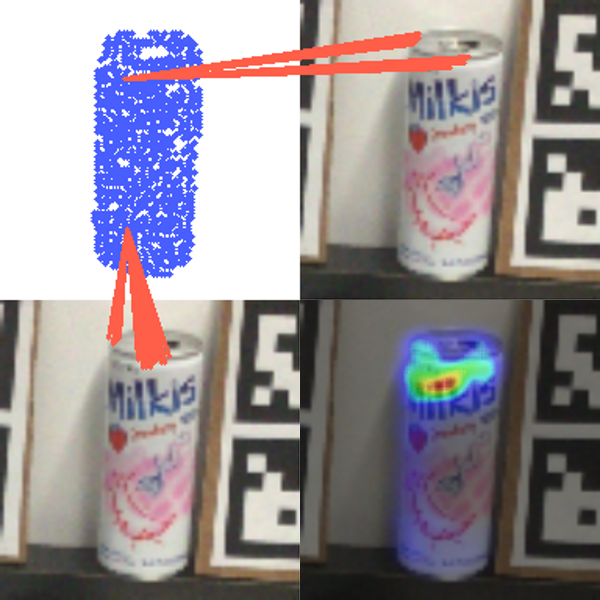}
    \end{minipage}
    \vspace{2pt}
    \vspace{-2pt}
    \caption{Visualization of learned feature relations from instance point clouds to RGB images. The top-ranking relations (i.e., red lines) are computed from the learned network parameters (i.e., transformers in IRN). In each $2\times2$ cell, red lines indicate the top-20 relations for a 3D point, and the heatmap at the lower right corner illustrates the overall relation distribution of instance point clouds.}
    \label{fig:relation1}
    \vspace{-8pt}
\end{figure*}

\subsection{Comparison with State-of-the-Art Methods}\label{subsec:sota_experiment}
We compare with three state-of-the-art methods: NOCS \cite{wang2019normalized}, Shape Prior Deformation~(SPD) \cite{tian2020shape} and Canonical Shape Space~(CASS) \cite{chen2020learning}. 
To our knowledge, these are all the popular methods that address the challenging category-level 6D pose estimation task in current literature.
Table~\ref{table:stoa} and Figure~\ref{fig:qua_com} present the quantitative and qualitative results.

\textbf{CAMERA25.} On this benchmark dataset, our method significantly outperforms all previous state-of-the-arts approaches across all metrics. Notably, on the two most strict metrics, we achieve $88.0\%$ on $3D_{75}$ and $72.0\%$ on $5^{\circ}2 cm$, which are $18.5\%$ and $39.7\%$ higher than NOCS~\cite{wang2019normalized}, and $4.9\%$ and $17.7\%$ higher than SPD~\cite{tian2020shape}. Such large margins exceeding the current methods are attributed to our relation-enhanced representation learning and recurrent refinement of NOCS reconstruction model. On those relatively flexible metrics of $3D_{50}$ and  $10^{\circ}2 cm$, our mean average precision (mAP) reaches a high value of $93.8\%$ and $87.7\%$, indicating the promising potential for practical use.


\textbf{REAL275.} The dataset of REAL275 is much more challenging than CAMERA25 given the real-world complications and the limited amount of training data. Only $3$ object instances of each category are given for training, and 3 new instances are given for testing.
Thus, in consistency with~\cite{tian2020shape,wang2019normalized}, we randomly select data from CAMERA25 and REAL275 at a ratio of 3:1 and train the network using this hybrid dataset. Under this setting, our method achieves a mAP of $55.9\%$ for 3D IoU at $75\%$, and a mAP of $27.8\%$ for pose error within $5^{\circ}2$ cm. These performances are $25.9\%$ and $20.6\%$ higher than NOCS~\cite{wang2019normalized}, and $2.7\%$ and $8.5\%$ higher than SPD~\cite{tian2020shape}. We also compare our results with the accuracy value reported in CASS~\cite{chen2020learning}, and our method outperforms it by $10.8\%$ on $5^{\circ}5cm$, as well as by 1.6\% and by $2.8$ on the less strict metrics of $3D_{50}$ and $10^{\circ}5cm$.

\textbf{NOCS Model Reconstruction.}
To evaluate the quality of our reconstructed NOCS model regarding the shape of instance point cloud,
we report the Chamfer Distance(CD) between our reconstructed result and the ground truth model.
Table \ref{table:reconstruction} compares our performance on the CD metric with SPD~\cite{tian2020shape}.
It is observed that our method can consistently improve the accuracy of reconstructions over the state-of-the-art SPD~\cite{tian2020shape} across all categories on CAMERA25 and on five out of the six categories on REAL275. 
These analysis validate our superior quality of reconstructed NOCS model particularly with a precise shape which is important for accurate category-level pose estimation.



\subsection{Ablation Studies}\label{subsec:ablation}
To investigate the properties of our method, we evaluate its key model components on CAMEAR25 and REAL275. 

\textbf{Cascaded Relation.} In Table~\ref{table:relation}, we quantitatively evaluated our cascaded relation network with different settings. Specifically, we test three well-known relational structures: MLP~\cite{santoro2017simple}, Non-Local~\cite{wang2018non}, and Transformer~\cite{devlin2018bert}. According to the results in~Table~\ref{table:relation}, we find that no matter which structure is adopted, the pose estimation results get consistent improvement over the baseline. Moreover, we observe that the Transformer-based structure achieves the best pose estimation results on both datasets, which indicates the superiority of recent Transformer modules in modelling long-range and/or cross-modal feature relations. In addition, we further remove the IRN and CRN individually and observe the change of pose estimation results (see the row of ``- / T" and ``T / -" respectively). The obtained results demonstrate that both IRN and CRN are important in our network with balanced contribution to the performance. Cascading them together can take full advantages of the inherent feature relations for 6D object pose estimation. 


\textbf{Intuitive Visualizations of Learned Relations.}
We visualize the relations from instance point clouds to RGB image that are learnd by our relation network for interpretable understandings of what have been captured in the networks. As presented in Figure~\ref{fig:relation1}, the top-20 relations and heatmaps would attend more to the boundary regions of the objects, which are very informative for estimating object 6D pose.



\textbf{Recurrent Reconstruction.} In Table~\ref{table:recurrent}, we investigate our recurrent reconstruction network on different numbers of recurrent steps. ``0" denotes the baseline network without any recurrent refinement. From the mAP measurements, we find that gradually increasing the number of recurrent step progressively improves the pose estimation outcomes. The pose accuracy consistently increases in the first two iterations and gets relatively stable after that. With a 2-step recurrent, for the most strict metric $5^{\circ}2cm$, our recurrent network improves the mAP from $59.7\%$ to $64.4\%$ on CAMERA25 data and from $20.2\%$ to $24.2\%$ on REAL275 data. These results demonstrate that harnessing our proposed recurrent reconstruction network is beneficial for boosting pose estimation accuracy.



%% file: sections/conclusion.tex

\section{Conclusion}

We have presented a novel cascaded relation and recurrent reconstruction framework for category-level 6D object pose and size estimation. Our approach models the relations of RGB image, point cloud, and shape prior through a devised cascaded relation network, which enables our network to learn more representative instance features and overcome shape variations of different instances. The NOCS canonical object shape is reconstructed from a coarse-to-fine manner via a designed recurrent reconstruction network. Extensive experimental results present dramatic performance improvement over existing methods, setting new state-the-the-art results on both benchmarks.
Our method has promising potential to be applied for downstream applications such as robotic manipulation of objects in real environments.


%% file: sections/acknowledge.tex
\noindent\textbf{Acknowledgement.}
The work was supported by the Hong Kong Centre for Logistics Robotics.